\documentclass[runningheads]{llncs}
\usepackage[utf8]{inputenc}
\usepackage{amsmath}
\usepackage{amsfonts}
\usepackage{algpseudocode}
\usepackage{graphicx}
\usepackage{url}
\usepackage[ruled]{algorithm2e}

\title{ChatGPT is not all you need. A State of the Art Review of large Generative AI models}
\titlerunning{State of the Art of Generative AI}
\author{Roberto Gozalo-Brizuela, Eduardo C. Garrido-Merchán}
\institute{Quantitative Methods Department, Universidad Pontificia Comillas, Madrid, Spain
\email{201905616@alu.comillas.edu, ecgarrido@icade.comillas.edu}}

\begin{document}

\maketitle

\begin{abstract}
During the last two years there has been a plethora of large generative models such as ChatGPT or Stable Diffusion that have been published. Concretely, these models are able to perform tasks such as being a general question and answering system or automatically creating artistic images that are revolutionizing several sectors. Consequently, the implications that these generative models have in the industry and society are enormous, as several job positions may be transformed. For example, Generative AI is capable of transforming effectively and creatively texts to images, like the DALLE-2 model; text to 3D images, like the Dreamfusion model; images to text, like the Flamingo model; texts to video, like the Phenaki model; texts to audio, like the AudioLM model; texts to other texts, like ChatGPT; texts to code, like the Codex model; texts to scientific texts, like the Galactica model or even create algorithms like AlphaTensor. This work consists on an attempt to describe in a concise way the main models are sectors that are affected by generative AI and to provide a taxonomy of the main generative models published recently.   
\end{abstract}

\section{Introduction}

Generative AI refers to artificial intelligence that can generate novel content, rather than simply analyzing or acting on existing data like expert systems \cite{murphy2022probabilistic}. In particular, expert systems contained knowledge bases and an inference engine that generated content via an if-else rule database. However, modern generative artificial intelligence contain a discriminator or transformer model trained on a corpus or dataset that is able to map the input information into a latent high-dimensional space and a generator model, that is able to generate an stochastic behaviour creating novel content in every new trial even from the same prompts as an input, performing unsupervised, semi-supervised or supervised learning, depending on the particular methodology. Regarding the created content by the model, generative artificial intelligence models are different from predictive machine learning systems, that merely perform a discrimination behaviour, solving classification or regression problems. In particular, these models are able to discriminate information and generate information of the transformed input information, or prompt. 

The key aspect about generative models is that their architecture and the data that they have been fed is enormous. For example, it is possible now to estimate the parameters of the model by feeding it the contents of the whole Wikipedia, Github, social networks, Google images and more. Despite being fed with an enormous size of data, thanks to the rise of computing we can design deep neural networks \cite{lecun2015deep}, transformers \cite{lin2022survey} and other models such as generative adversarial networks \cite{creswell2018generative} or variational autoencoders \cite{murphy2022probabilistic} whose capacity is able to model the complexity of the data, without suffering from underfitting. As they are able to modelize the high-dimensional probability distribution of language or photos of a concrete or general domain, if they are complemented by generative models that map the latent high-dimensional semantic space of language of photos to a multimedia representation of text, audio or video we can map any input format like texts to any output format like video. In this sense, applications of this technology are endless, in the sense that we can train a model to generate genuine different multimedia formats as video, audio or text from different multimedia input formats, as for example, text. 

We believe that it is necessary to provide a state-of-the-art review on the most popular generative AI models as they are revolutionizing several industries like the art industry \cite{anantrasirichai2021artificial} or universities \cite{kandlhofer2016artificial,susnjak2022chatgpt}. As models are now able to generate genuine artistic content or large texts answering a prompt, these two industries and other ones that we will detail throughout this manuscript will need to readapt their activity to continue providing value. In this sense, generative AI models will not replace humans but enhance our content, being an inspiration for artists or improving the content generated by professors. In order to provide information for a professional working in any industry that can be benefited by these models we have made the organization of the paper as the following one. First, we will provide a taxonomy of the main generative models that have appeared in the industry. Then, the following sections will analyze each of the categories of the taxonomy. Finally, we finish the manuscript with a conclusions and further work section. We do not study the technical aspects of every model, such as transformers in detail as our purpose in this review is on the applications of the models and the content that they generative but not on how they work. For a detailed explanation of deep learning models and generative models we recommend other references \cite{lecun2015deep,murphy2022probabilistic}.   

\section{A Taxonomy of Generative AI models}
\begin{figure}[htb!]
    \centering
    \includegraphics[width = 0.9\textwidth]{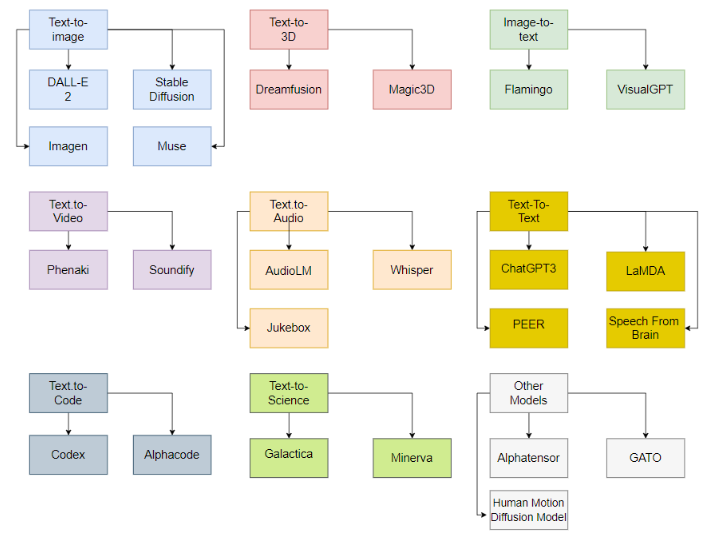}
\caption{A taxonomy of the most popular generative AI models that have recently appeared classified according to their input and generated formats.}
    \label{fig:taxonomy}
\end{figure}

\begin{figure}[htb!]
    \centering
    \includegraphics[width = 0.99\textwidth]{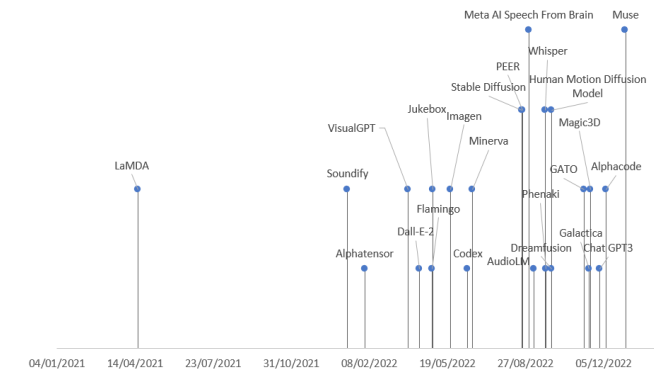}
\caption{Covered models by date of release. All models were released during 2022 except LaMDA, which was released in 2021 and Muse, in 2023.}
    \label{fig:dates}
\end{figure}

Before analyzing each model in detail, we have tried to organize the current generative artificial models into a taxonomy whose categories represent the main mappings between each multimedia input and output type of data. The result is the one that we have illustrated in Figure \ref{fig:taxonomy}. We have discovered a total of 9 categories, where each of the models that appear in Figure \ref{fig:taxonomy} will be described in detail in the following section.   
Each of the covered models has been published recently, as we illustrate in Figure \ref{fig:dates}, as our main concern in this manuscript is to describe the latest advances in generative AI models. 

\begin{figure}[htb!]
    \centering
    \includegraphics[width = 0.8\textwidth]{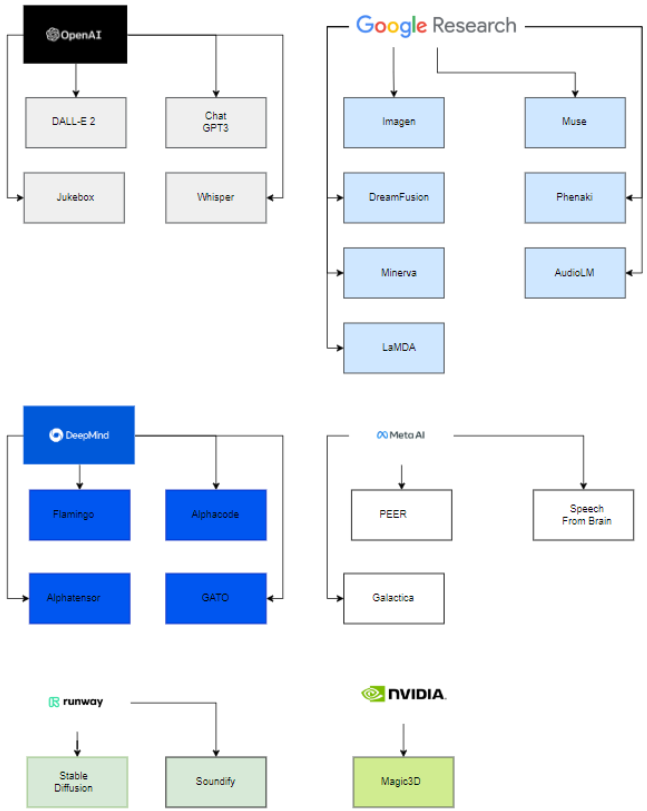}
\caption{Models by developer. In terms of major companies participating in startups, note that Microsoft invested 1 billion dollars in OpenAI and helps them with the development of models. As well, note that Google acquired Deepmind in 2014. In terms of universities, note that VisualGPT was developed by KAUST, Carnegie Mellon University and Nanyang Technological University and that the Human Motion Diffusion Model was developed by Tel Aviv University, Israel. As well, other projects are developed by a company in collaboration with a university. Concretely, this is the case for Stable Diffsion (Runway, Stability AI and LMU MUNICH), Soundify (Runway and Carnegie Mellon University) and DreamFusion (Google and UC Berkeley)}
    \label{fig:companies}
\end{figure}

Interestingly, only six organizations are behind the deployment of these models, as we illustrate in Figure \ref{fig:companies}. The main reason behind this fact is that in order to be able to estimate the parameters of these models, it is mandatory to have an enormous computation power and a highly skilled and experienced team in data science and data engineering. Consequently, only the companies shown on Figure \ref{fig:companies}, with the help of acquired startups and collaborations with academia, have been able to successfully deploy generative artificial intelligence models.      

Now that we have provided and analyzed the latest generative artificial intelligence models, the following section will cover each of the categories of the taxonomy presented in Figure \ref{fig:taxonomy} in detail.

\section{Generative AI models categories}
In this section we will cover in detail the nine categories described in Figure \ref{fig:taxonomy} of the previous section. For every category, we illustrate the details of the models shown in Figure \ref{fig:taxonomy}.  
\subsection{Text-to-image models}
We begin the review by considering the models whose input is a text prompt and whose output is an image. 
\subsubsection{DALL$\cdot$E 2}
\begin{figure}[htb!]
    \centering
    \includegraphics[width = 0.3\textwidth]{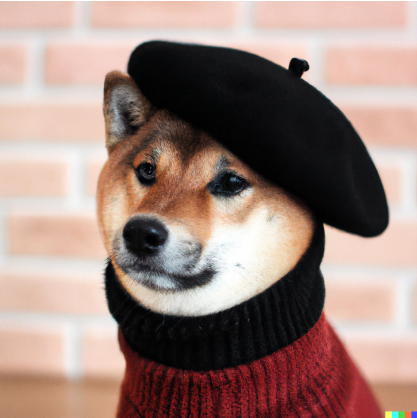}
\caption{Image generated from the prompt "A shiba inu wearing a beret and black turtleneck".}
    \label{fig:dalle2}
\end{figure}

:
DALL$\cdot$E 2, created by OpenAI, is able to generate original, genuine and realistic images and art from a prompt consisting on a text description \cite{daras2022discovering}. Luckily, it is possible to use the OPENAI API to get access to this model. In particular, DALL$\cdot$E 2 manages to combine concepts, attributes and different styles. In order to do so, it uses the CLIP neural network. CLIP (Contrastive Language-Image Pre-Training) is a neural network trained on a variety of (image, text) pairs \cite{radford2021learning}. Using CLIP, that can be instructed in natural language to predict the most relevant text snippet, given an image, the model has recently merged as a successful representation learner for images. Concretely, CLIP embeddings have several desirable properties: they are robust to image distribution shift, have impressive zero-shot capabilities and have been fine-tuned to achieve state-of-the-art results. In order to obtain a full generative model of images, the CLIP image embedding decoder module is combined with a prior model, which generates possible CLIP image embeddings from a given text caption. We illustrate an image generated from a prompt in Figure \ref{fig:dalle2}

\subsubsection{IMAGEN}
\begin{figure}[htb!]
    \centering
    \includegraphics[width = 0.4\textwidth]{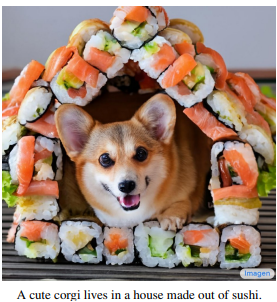}
\caption{Image generated from the prompt "A cute corgi lives in a house made out of sushi".}
    \label{fig:image_imagen}
\end{figure}

: Imagen is a text-to-image diffusion model \cite{kingma2021variational} consisting on large transformer language models. Critically, the main discovery observed with this model made is that large language models, pre-trained on text-only corpora, are very effective at encoding text for image synthesis \cite{saharia2022photorealistic}. Precisely, using Imagen, it has been found out that increasing the size of the language model boosts both sample fidelity and image-text alignment much more than increasing the size of the image diffusion model. The model was created by Google and the API can be found in their web page. For the evaluation of their model, Google created Drawbench, a set of 200 prompts that support the evaluation and comparison of text-to-image models. Most concretely, the model is based on a pretrained text encoder (like BERT \cite{devlin2018bert}) that performs a mapping from text to a sequence of word embeddings and a cascade of conditional diffusion models that map these embeddings to images of increasing resolutions. We show an image generated from a prompt in Figure \ref{fig:image_imagen}.

\subsubsection{Stable Diffusion}: Stable Diffusion is a latent-diffusion model that is open-source and has been developed by the CompVis group at LMU Munich. The main difference of this model with respect to the other ones is the use of a latent diffusion model and that it performs  image modification as it can perform operations in its latent space. For Stable Diffusion, we can use the API via their website. More concretely, Stable Diffusion consists of two parts: the text encoder and the image generator [17]. The image information creator works completely in the latent space. This property makes it faster than previous diffusion models that worked in a pixel space. We illustrate a Stable Diffusion image example in Figure \ref{fig:stable_diffusion}.

\begin{figure}[htb!]
    \centering
    \includegraphics[width = 0.4\textwidth]{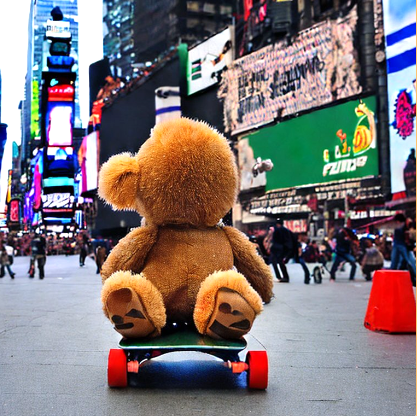}
\caption{Image generated from the prompt "A cute corgi lives in a house made out of sushi".}
    \label{fig:stable_diffusion}
\end{figure}

\subsubsection{Muse}: This model is a Text-to-image transformer model that achieves state-of-the-art image generation while being more efficient than diffusion or autoregressive models \cite{chang2023muse}. Concretely, it is trained on a masked modelling task in discrete token space. Consequently, it is more efficient because of the use of discrete tokens and requiring fewer sampling iterations. Compared to Parti, a autoregressive model, Muse is more efficient because of parallel decoding. Muse is 10x faster at inference time than Imagen-3B or Parti-3B and 3x faster than Stable Diffusion v 1.4. Muse is also faster than than Stable Diffusion in spite of both models working in the latent space of a VQGAN. We append a comparison of the generated images by DALL$\cdot$E 2, IMAGEN and Muse in Figure \ref{fig:image_comparison}.

\begin{figure}[htb!]
    \centering
    \includegraphics[width = 0.99\textwidth]{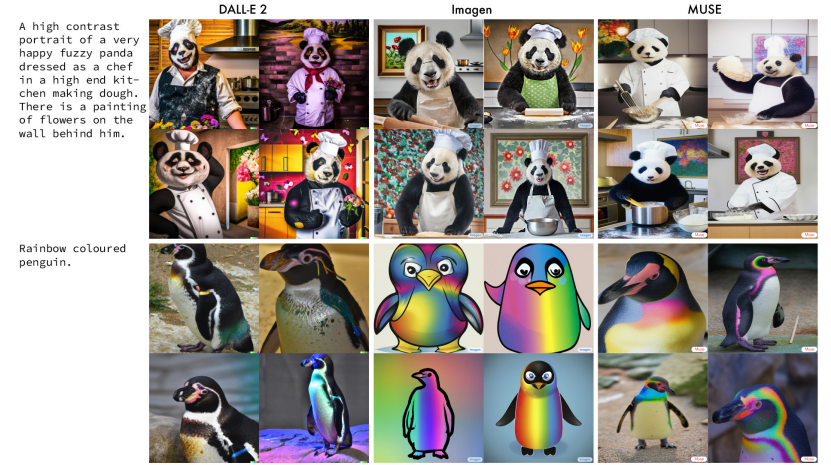}
\caption{Comparison of generated images by the DALL$\cdot$E 2, IMAGEN and Muse models with respect to the prompts that appear in the column of the left. The first column of images contains the results generated by DALL$\cdot$E 2, the second the results obtained with IMAGEN and the third the images created by Muse.}
    \label{fig:stable_diffusion}
\end{figure}

\subsection{Text-to-3D models}
The models that have been described in the previous section deal with the mapping of text prompts to 2D images. However, for some industries like gaming, it is necessary to generate 3D images. In this section, we briefly describe two text-to-3D models: Dreamfusion and Magic3D.

\subsubsection{Dreamfusion}
\begin{figure}[htb!]
    \centering
    \includegraphics[width = 0.99\textwidth]{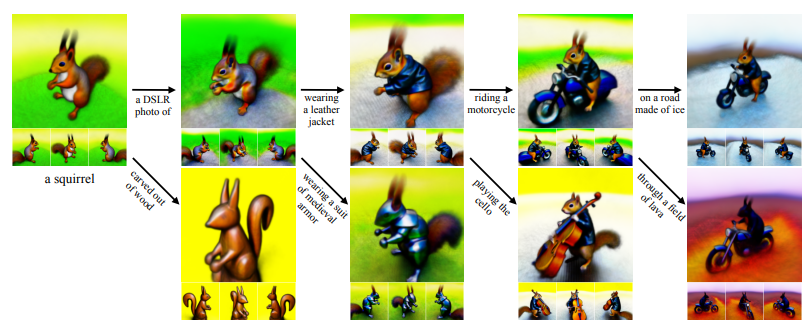}
\caption{A 3D squirrel generated by Dreamfusion is shown at the left. Then, the other images contain the modifications generated to the squirrel with text prompts like "wearing a jacket".}
    \label{fig:dreamfusion}
\end{figure}

: DreamFusion is a text-to-3D model developed by Google Research that uses a pretrained 2D text-to-image diffusion model to perform text-to-3D synthesis \cite{poole2022dreamfusion}. In particular, Dreamfusion replaces previous CLIP techniques with a loss derived from distillation of a 2D diffusion model. Concretely, the diffusion model can be used as a loss within a generic continuous optimization problem to generate samples. Critically, sampling in parameter space is much harder than in pixels as we want to create 3D models that look like good images when rendered from random angles. To solve the issue, this model uses a differentiable generator. Other approaches are focused on sampling pixels, however, this model instead focuses on creating 3D models that look like good images when rendered from random angles. We illustrate in Figure \ref{fig:dreamfusion} an example of an image created by Dreamfusion from one particular angle along with all the variations that can be generated from additional text prompts. In order to see the full animated image, we recommend to visit the web page of Dreamfusion.

\subsubsection{Magic3D}
\begin{figure}[htb!]
    \centering
    \includegraphics[width = 0.99\textwidth]{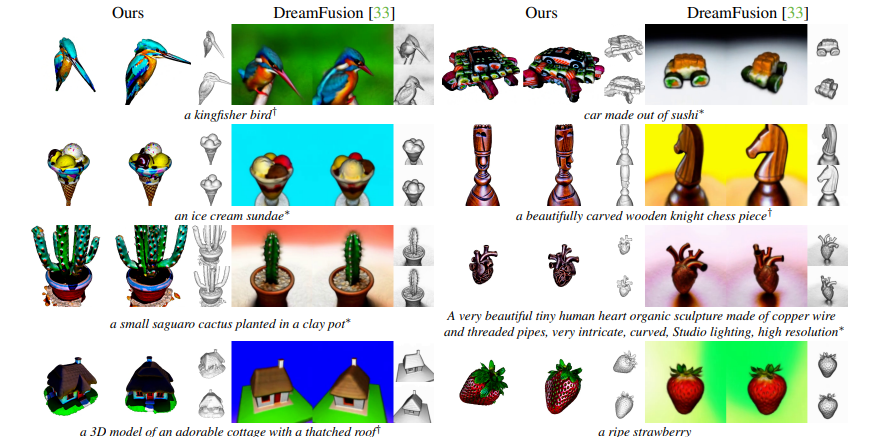}
\caption{3D Images generated by Magic3D and Dreamfusion, where "Ours" refer to Magic3D. We can see a total of 8 text prompts and the images that both models generate from that prompts.}
    \label{fig:magic3d}
\end{figure}

: This model is a text to 3D model made by NVIDIA Corporation. While the Dreamfusion model achieves remarkable results, the method has two problems: mainly, the long processing time and the low-quality of the generated images. However, these problems are addressed by Magic3D using a two-stage optimization framework \cite{lin2022magic3d}. Firstly, Magic3D builds a low-resolution diffusion prior and, then, it accelerates with a sparse 3D hash grid structure. Using that, a textured 3D mesh model is furthered optimized with an efficient differentiable render. Comparatively, regarding human evaluation, the model achieves better results, as 61.7\% prefer this model to DreamFusion. As we can see in Figure \ref{fig:magic3d}, Magic3D achieves much higher quality 3D shapes in both geometry and texture compared to DreamFusion. 

\subsection{Image-to-Text models}
Sometimes, it is also useful to obtain a text that describes an image, that is precisely the inverse mapping to the one that has been analyzed in the previous subsections. In this section, we analyze two models that perform this task, along with others: Flamingo and VisualGPT.

\subsubsection{Flamingo}
\begin{figure}[htb!]
    \centering
    \includegraphics[width = 0.99\textwidth]{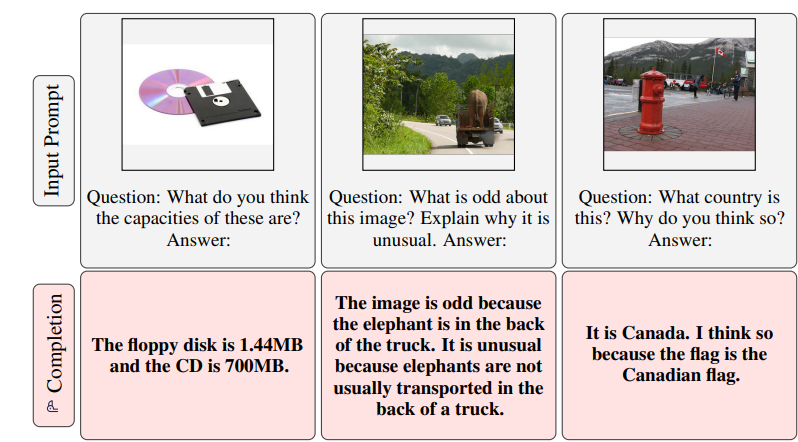}
\caption{Input prompts that contain images and text and output generated text respones from Flamingo. Every column contains a single example where we can see how Flamingo answers the question using the image from the text.}
    \label{fig:flamingo}
\end{figure}

: A Visual Language Model created by Deepmind using few shot learning on a wide range of open-ended vision and language tasks, simply by being prompted with a few input/output examples \cite{alayrac2022flamingo}. Concretely, the input of Flamingo contains visually conditioned autoregressive text generation models able to ingest a sequence of text tokens interleaved with images and/or videos and produce text as output. A query is made to the model along with a photo or a video and the model answers with a text answer. Some examples can be observed in Figure \ref{fig:flamingo}. Flamingo models take advantage of two complementary models: a vision model that analyzes visual scenes and a large language model which performs a basic form of reasoning. The language model is trained on a large amount of text data.

\subsubsection{VisualGPT}
\begin{figure}[htb!]
    \centering
    \includegraphics[width = 0.99\textwidth]{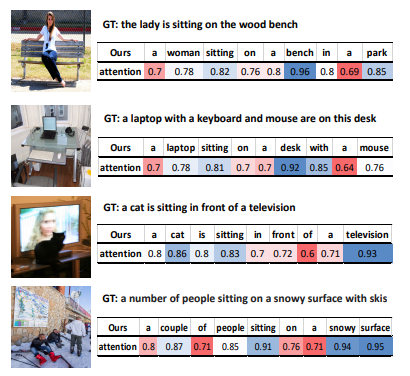}
\caption{Three examples of text prompts generated by the images shown on the left. We also show the attention scores that the model assign to every word of the texts. In the third image, we can see for example how the most discriminative information about the image is the word "cat" and "television".}
    \label{fig:visualgpt}
\end{figure}

: VisualGPT is an image captioning model made by OpenAI \cite{chen2022visualgpt}. Concretely, VisualGPT leverages knowledge from the pretrained language model GPT-2 \cite{budzianowski2019hello}. In order to bridge the semantic gap between different modalities, a novel encoder-decoder attention mechanism \cite{vaswani2017attention} is designed with an unsaturated rectified gating function. Critically, the biggest advantage of this model is that it does not need for as much data as other image-to-text models. In particular, improving data efficiency in image captioning networks would enable quick data curation, description of rare objects, and applications in specialized domains. Most interestingly, the API of this model can be found on GitHub. We include three examples of text prompts generated by the model with respect to three images fed to the model in Figure \ref{fig:visualgpt}. 

\subsection{Text-to-Video models}
As we have seen in the previous subsections, it is now possible to generate images from text. Consequently, the next logical step is to generate videos, that are sequences of images, from texts. In this section, we provide information about two models that are able to perform this task: Phenaki and Soundify.

\subsubsection{Phenaki}: This model has been made by Google Research, and it is capable of performing realistic video synthesis, given a sequence of textual prompts \cite{villegas2022phenaki}. Most interestingly, we can get access to the API of the model from GitHub. In particular, Phenaki is the first model that can generate videos from open domain time variable prompts. To address data issues, it performs joint training on a large image-text pairs dataset as well as a smaller number of video-text examples can result in generalization beyond what is available in the video datasets. This is mainly due to image-text datasets having billions of inputs while text-video datasets are much smaller. As well, limitations come from computational capabilities for videos of variable length. 

\begin{figure}[htb!]
    \centering
    \includegraphics[width = 0.99\textwidth]{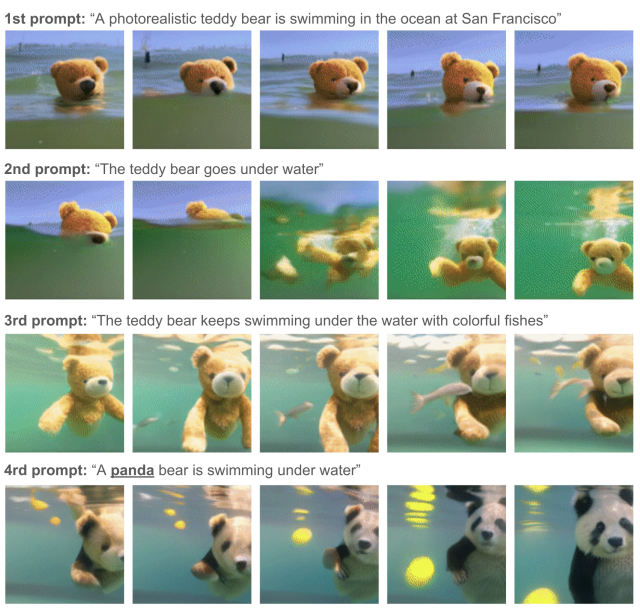}
\caption{Sequence of images created by the Phenaki model given four different prompts.}
    \label{fig:phenaki}
\end{figure}

\begin{figure}[htb!]
    \centering
    \includegraphics[width = 0.99\textwidth]{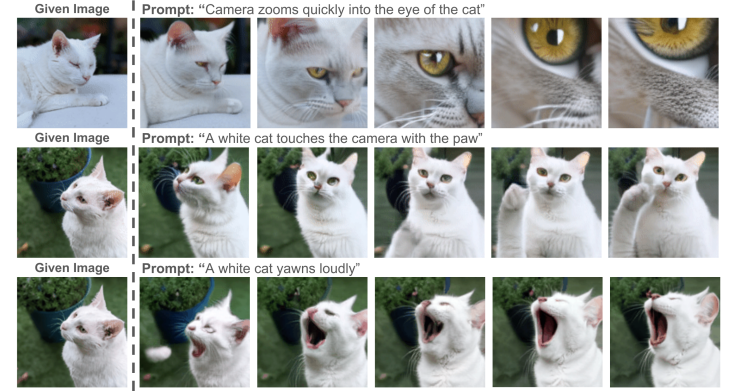}
\caption{Sequences of images created by the Phenaki model given an image and the prompt. We can see how the model is able to manipulate the given image according to the text prompt.}
    \label{fig:phenaki2}
\end{figure}

The model has three parts: the C-ViViT encoder, the training transformer and the video generator. The encoder gets a compressed representation of videos. First tokens are transformed into embeddings. This is followed by the temporal transformer, then the spatial transformer. After the output of the spatial transformer, they apply a single linear projection without activation to map the tokens back to pixel space. Consequently,  the model generates temporally coherent and diverse videos conditioned on open domain prompts even when the prompt is a new composition of concepts. The videos can be minutes long, while the model is trained on 1.4 second videos. Below we show in Figure \ref{fig:phenaki} and in Figure \ref{fig:phenaki2} some examples of the creation of a video through a series of text prompts and from a series of text prompts and an image. 

\subsubsection{Soundify}: In video editing, sound in half of the story. But, for professional video editing, the problems come from finding suitable sounds, aligning sounds, video and tuning parameters \cite{lin2021soundify}. In order to solve this issue, Soundify is a system developed by Runway that matches sound effects to video. This system uses quality sound effects libraries and CLIP (a neural network with zero-shot image classification capabilities cited before). Concretely, the system has three parts: classification, synchronization, and mix. The classification matches effects to a video by classifying sound emitters within. To reduce the distinct sound emitters, the video is split based on absolute color histogram distances. In the synchronization part, intervals are identified comparing effects label with each frame and pinpointing consecutive matches above a threshold. In the mix part, effects are split into around one-second chunks. Critically, chunks are stitched via crossfades.

\subsection{Text-to-Audio models}
As we have seen in the previous subsection, images are not the only important non-structured data format. For videos, for music and in lots of contexts, audio can be critical. Consequently, we analyze in this subsection three models whose input information is text and whose output information is audio.

\subsubsection{AudioLM}: This model has been made by Google for high-quality audio generation with long-term consistency. In particular, AudioLM maps the input audio into a sequence of discrete tokens and casts audio generation as language modeling task in this representation space \cite{borsos2022audiolm}. By training on large corpora of raw audio waveforms, AudioLM learns to generate natural and coherent continuations given short prompts. The approach can be extended beyond speech by generating coherent piano music continuations, despite being trained without any symbolic representation of music. As with the other models, the API can be found through GitHub. Audio signals involve multiple scales of abstractions. When it comes to audio synthesis, multiple scales make achieving high audio quality while displaying consistency very challenging. This gets achieved by this model by combining recent advances in neural audio compression, self-supervised representation learning and language modelling.

In terms of subjective evaluation, raters were asked to listen to a sample of 10 seconds and decide whether it is human speech or a synthetic continuation. Based on 1000 ratings collected, the rate is 51.2\%, which is not statistically significant from assigning labels at random. This tells us that humans cannot differentiate between synthetic and real samples.

\subsubsection{Jukebox}: This is a model, developed by OpenAI, that generates music with singing in the raw audio domain \cite{dhariwal2020jukebox}. Once again, its API can be found in GitHub. Previously, earlier models in the text-to-music genre generated music symbolically in the form of a pianoroll which specifies timing, pitch and velocity. The challenging aspect is the non-symbolic approach where music is tried to be produced directly as a piece of audio. In fact, the space of raw audio is extremely high dimensional which makes the problem very challenging. Consequently, the key issue is that modelling that raw audio produces long-range dependencies, making it computationally challenging to learn the high-level semantics of music. 

In order to solve this issue, this model tries to solve it by means of a hierarchical VQ-VAE architecture to compress audio into a discrete space \cite{ding2019group}, with a loss function designed to retain the most amount of information. This model produces songs from very different genres such as rock, hip-hop and jazz. However, the model is just limited to English songs. Concretely, its dataset for training is from 1.2 million songs from LyricWiki. The VQ-VAE has 5 billion parameters and is trained on 9-second audio clips for 3 days.

\subsubsection{Whisper}: This model is an Audio-to-Text converter developed by OpenAI. It achieves several tasks in this field: multi-lingual speech recognition, translation and language identification \cite{radford2022robust}. As in previous cases, its API can be found in the GitHub website. The goal of a speech recognition system should be to work reliably out of the box in a broad range of environments without requiring supervised fine-tuning of a decoder for every deployment distribution. This is hard because of the lack of a high-quality pre-trained decoder.

Concretely, this model is trained on 680,000 hours of labeled audio data. This data is collected from the internet, which results in a very diverse dataset covering a broad distribution of audio from many different environments, recordings setups, speakers and languages. The model makes sure that the dataset is only from human voice as machine learning voice would impair the model. Files are broken in 30 second segments paired with the subset of the transcript that occurs within that time segment. 

The model has an encoder-deccoder transformer, as this architecture has been validated to scale reliably. We can observe the model’s architecture characteristics through the figure below. We can see the different types of data and the learning sequence.

\subsection{Text-to-Text models}
The previous models all convert a non-structured data type into another one. But, regarding text, it is very useful to convert text into another text in order to satisfy tasks as general question and answering. The following four models treat text and also output texts to satisfy different needs. 

\subsubsection{ChatGPT}
\begin{figure}[htb!]
    \centering
    \includegraphics[width = 0.99\textwidth]{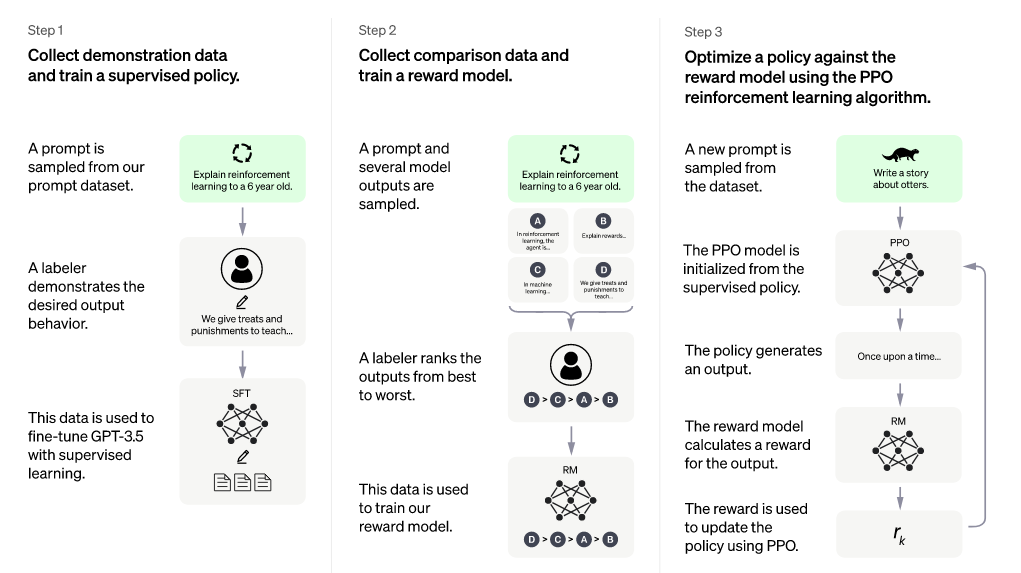}
\caption{Training steps of ChatGPT, combining supervised learning with reinforcement learning.}
    \label{fig:chatgpt}
\end{figure}

: The popular ChatGPT is a model by OpenAI which interacts in a conversational way. As it is widely known, the model answers follow-up questions, challenges incorrect premises and reject inappropriate requests. More concretely, the algorithm behind ChatGPT is based on a transformer. However, the training is made through Reinforcement Learning for Human Feedback. In particular, an initial model is trained using supervised fine-tuning: human AI trainers would provide conversations in which they played both sides, the user and an AI assistant. Then, those people would be given the model-written responses to help them compose their response. This dataset was mixed to that of InstructGPT \cite{bhavya2022analogy}, which was transformed into a dialogue format. A demo can be found in their website and the API may also be found in OpenAI’s website. We summarize the main steps of ChatGPT training in Figure \ref{fig:chatgpt}, available in the ChatGPT demo's website. Finally, ChatGPT is also able to generate code and simple mathematics.  

\subsubsection{LaMDA}
\begin{figure}[htb!]
    \centering
    \includegraphics[width = 0.99\textwidth]{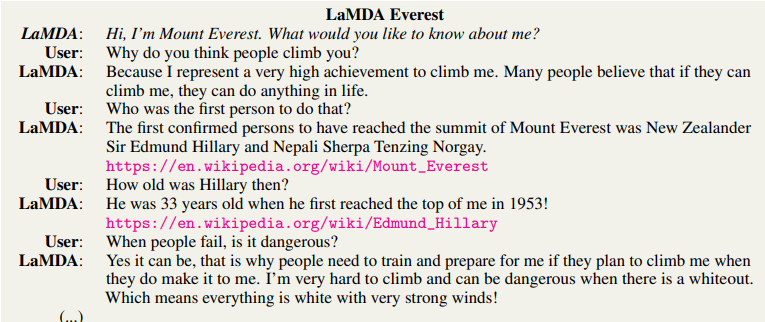}
\caption{Example of a dialog made with LaMDA.}
    \label{fig:lamda}
\end{figure}

: LaMDA is a language model for dialog applications \cite{thoppilan2022lamda}. Unlike most other language models, LaMDA was trained on dialogue. It is a family of transformer-based neural language models specialized for dialog which have up to 137B parameters and are pre-trained on 1.56T words of public dialog data and web text. Fine-tuning can enable for safety and factual grounding of the model. Only 0.001\% of training data was used for fine-tuning, which is a great achievement of the model. In particular, dialog modes take advantage of Transformers’ ability to present long-term dependencies in text. Concretely, they are generally very well-suited for model scaling. Consequently, LaMDA makes use of a single model to perform multiple tasks: it generates several responses, which are filtered for safety, grounded on an external knowledge source and re-ranked to find the highest-quality response. We illustrate in Figure \ref{fig:lamda} an example of a dialog with the model.

\subsubsection{PEER}: Collaborative language model developed by Meta AI research trained on edit histories to cover the entire writing process \cite{schick2022peer}. It is based on four steps: Plan, Edit, Explain and Repeat. These steps are repeated until the text is in a satisfactory state that requires no further updates. The model allow to decompose the task of writing a paper into multiple easier subtasks. As well, the model allows humans to intervene at any time and steer the model in any direction.

It is mainly trained on Wikipedia edit histories. The approach is a self-training, using models to infill missing data and then train other models on this synthetic data. The downside of this comes from comments being very noisy and a lack of citations, which tries to be compensated by a retrieval system which does not always work. The framework is based on an iterative process. The entire process of formulating a plan, collecting documents, performing an edit and explaining it can be repeated multiple times until arriving at a sequence of texts. For the training, a DeepSpeed transformer is used.

\subsubsection{Meta AI Speech from Brain}: Model developed by Meta AI to help people unable to communicate through speech, typing or gestures \cite{defossez2022decoding}. Previous techniques relied on invasive brain-recording techniques which require neurosurgical interventions. This model tries to decode language directly from noninvasive brain recordings. This would provide a safer, more scalable solution that could benefit many more people. The challenge with this proposed method come from noise and differences in each person’s brain and where the sensors are placed.

A deep learning model is trained with contrastive learning and used to maximally align noninvasive brain recordings and speech sounds. A self-supervised learning model called wave2vec 2.0. is used to identify the complex representations of speech in the brains of volunteers listening to audiobooks. The two noninvasive technologies used to measure neuronal activity are electroencephalography and magnetoencephalography.

Training data comes from four opensource datasets which represent 150 hours of recordings of 169 volunteers listening to audiobooks. EEG and MEG recordings are inserted into a brain model, which consists of a standard deep convolutional network with residual connections. These recordings are what comes from individuals’ brains. This model then has both a speech model for sound and a brain model for MEG data. 

Results show that several components of the algorithm were beneficial to decoding performance. As well, analysis shows that the algorithm improves as EEG and MEG recordings increase. This research shows that self-supervised trained AI can decode perveived speech despite noise and variability in that data. The biggest limitation of this research is that it focuses on speech perception, but the ultimate goal would be to extend this work to speech production.

\subsection{Text-to-Code models}
Although we have covered text-to-text models, not all texts follows the same syntax. An special type of text is code. In programming, it is essential to know how to convert an idea into code. In order to do so, Codex and Alphacode models help.

\subsubsection{Codex}: AI system created by OpenAI which translates text to code. It is a general-purpose programming model, as it can be applied to basically any programming task \cite{chen2021evaluating}. Programming can be broken down into two parts: breaking a problem down into simpler problems and mapping those problems into existing code (libraries, APIs, or functions) that already exist. The second part is the most time-barring part for programmers, and it is where Codex excels the most. The data collected for training was collected in May 2020 from public software repositories hosted on GitHub, containing 179GB of unique Python files under 1 MB. The model is fine-tuned from GPT-3, which already contains strong natural language representations. The demo and the API can be found in Open AI’s website. 

\subsubsection{Alphacode}: Other language models have demonstrated an impressive ability to generate code, but these systems still perform poorly when evaluated on more complex, unseen problems. However, Alphacode is a system for code generation for problems that require for deeper reasoning \cite{li2022competition}. Three components are key for this achievement: having an extensive dataset for training and evaluation, large and efficient transformer based architectures and a large-scale model sampling.

In terms of training, the model is firstly pre-trained  through GitHub repositories amounting to 715.1 GB of code. This is a much more extensive dataset than Codex’s pre training dataset. For the training to be better, a fine-tuning dataset is introduced from the Codeforces plataform. Through this platform, Codecontests are conducted, for the validation phase, in which we better the performance of the model. Regarding the transformer-based architecture, they use an encoder-decoder transformer architecture. Compared to decoder-only architectures commonly used, this architecture allows for a bidirectional description and extra flexibility. As well, they use a shallow encoder and a deep encoder to further the model’s efficiency. To reduce the cost of sampling, multi-query attention is used.

\subsection{Text-to-Science models}
Even scientific texts are being targeted by generative AI, as the Galactica and Minerva models have shown. Although there is a long way to manage success in this field, it is critical to study the first attempts towards automatic scientific text generation.

\subsubsection{Galactica}: Galactica is a new large model for automatically organizing science developed by Meta AI and Papers with Code. The main advantage of the model is the ability to train on it for multiple epochs without overfitting, where upstream and downstream performance improves with use of repeated tokens. The dataset design is critical to the approach as all of it is processed in a common markdown format to blend knowledge between sources. Citations are processed via a certain token that allows researchers to predict a citation given any input context. The capability of the model of predicting citations improves with scale and the model becomes better at the distribution of citations. In addition, the model can perform multi-modal tasks involving SMILES chemical formulas and protein sequences. Concretely, Galactica uses a transformer architecture in a decoder-only setup with GeLU activation for all model sizes.

\subsubsection{Minerva}: Language model capable of solving mathematical and scientific questions using step-by-step reasoning. Minerva has a very clear focus on the collection of training data for this purpose. It solves quantitative reasoning problems, makes models at scale and employs best-in-class inference techniques. Concretely, Minerva solves these problems by generating solutions step-by-step, this means including calculations and symbolic manipulation without having the need for external tools such a calculator. 

\subsection{Other models}
We would like to finish our review by covering additional models that do not fit any of the categories mentioned previously.

Alphatensor, created by the research company Deepmind, is a completely revolutionary model in the industry for its ability to discover new algorithms \cite{fawzi2022discovering}. In the published example, Alpha Tensor creates a more efficient algorithm for matrix multiplication, which is very important, as improving the efficiency of algorithms affects a lot of computations, from neural networks to scientific computing routines. 

The methodology is based on a deep reinforcement learning approach in which the agent, AlphaTensor is trained to play a single-player game where the objective is finding tensor decomposisitions within a finite factor space. At each step of the TensorGame, the player selects how to combine different entries of the matrices to multiply. A score is assigned based on the number of selected operations required to reach the correct multiplication result. To solve TensorGame, an agent, AlphaTensor was developed. AlphaTensor uses a specialized neural network architecture to exploit symmetries using synthetic training games.

GATO is a single generalist agent made by Deepmind. It works as a multi-modal, multi-task, multi-embodiment generalist policy \cite{reed2022generalist}. The same network with the same weights can carry very different capabilities from playing Atari, caption images, chatting, stacking blocks and many more. There are many benefits from using a single neural sequence model across all tasks. It reduces the need for hand crafting policy models with their own inductive biases. It increases the amount and diversity of training data. This general agent is successful at many tasks and can be adapted with little extra data to succeed at an even larger number of tasks. r training at the operating point of model scale that allows real-time control of real-world robots, currently around 1.2B parameters in the case of GATO.

Other published generative AI models are able to generate human motion \cite{tevet2022human} or, in the case of ChatBCG, slides using ChatGPT as a surrogate model.

\section{Conclusions and further work}
Through this paper, we can observe the capabilities which generative artificial intelligence has. We have seen a great deal of creativity as well as personalization in tasks such as text-to-image or in tasks such as text-to-audio. They also are accurate in text-to-science or text-to-code tasks. This can help economies in a major way as it can help optimize creative and non-creative tasks.

However, because of the way that they are constructed at the moment, these models face a number of limitations. In terms of dataset, finding data for some of the models found such as the text-to-science or the text-to-audio is very hard, making it very time-consuming to train the model. In particular, datasets and parameters have to be enormous, making it harder to train.  One of the biggest issues with models is trying solutions out of the problems in the dataset, with which models have more trouble solving. As well, in terms of computation, a lot of time and computation capacity is necessary in order to run them. Many days and advanced computers are needed in order to run the models. 

In addition, these models face bias from the data which needs to be controlled. Galactica model tries to control  this issue through a layer of no bias, but it still a major issue for Generative Artificial Intelligence.

With the Minerva model, we can see that the model knows the steps which it needs to take to solve an equation. This is groundbreaking as one of the biggest limitations with these models is that the models do not understand exactly what they are doing. Moreover, it’s still an industry starting; thus accuracy is still an issue. Text-to-video models for example are only represented by Phenaki because how hard it is to produce accurate videos. Text-to-science models do find some accuracy but that accuracy is still way behind to what it should be for professionals to actually rely on this technology on a day-to-day basis.

Furthermore, these models need to be constrained because of a lack of understanding of ethics.  Phenaki on its paper even acknowledges that a system like text-to-video can be used to create deep-fakes. Lastly, we are still in a phase where we are discovering what exactly the purpose of this intelligence will be. There has been articles comparing Google to ChatGPT3, which is totally inexact as ChatGPT3 does not update its information in real time. We should be aware about the limitations of these models to try and improve them in the following years.

\bibliographystyle{acm}

\bibliography{main}

\end{document}